\documentclass[runningheads]{llncs}
\pdfoutput=1
\usepackage{graphicx}
\usepackage{amsmath,amssymb} 
\usepackage{color}
\usepackage[width=122mm,left=12mm,paperwidth=146mm,height=193mm,top=12mm,paperheight=217mm]{geometry}
\usepackage{cite}
\usepackage{wrapfig}
\usepackage[hidelinks]{hyperref}
\usepackage{floatrow}

\newfloatcommand{captabbox}{table}[][\FBwidth]
\newfloatcommand{captabfig}{figure}[][\FBwidth]

\begin{document}
\pagestyle{headings}
\mainmatter

\title{Stacked Hourglass Networks for \\ Human Pose Estimation}

\titlerunning{Stacked Hourglass Networks for Human Pose Estimation}

\authorrunning{Newell et al.}

\author{Alejandro Newell, Kaiyu Yang, and Jia Deng}
\institute{University of Michigan, Ann Arbor\\
        \email{ \{alnewell,yangky,jiadeng\}@umich.edu}
}

\maketitle

\begin{abstract}
  This work introduces a novel convolutional network architecture for
  the task of human pose estimation. Features are processed across all
  scales and consolidated to best capture the various spatial
  relationships associated with the body. We show how repeated
  bottom-up, top-down processing used in conjunction with intermediate
  supervision is critical to improving the performance of the
  network. We refer to the architecture as a ``stacked hourglass''
  network based on the successive steps of pooling and upsampling that
  are done to produce a final set of predictions. State-of-the-art
  results are achieved on the FLIC and MPII benchmarks outcompeting
  all recent methods.

  \keywords{Human Pose Estimation}
\end{abstract}

\begin{figure}
\centering
\includegraphics[width=\textwidth, clip]{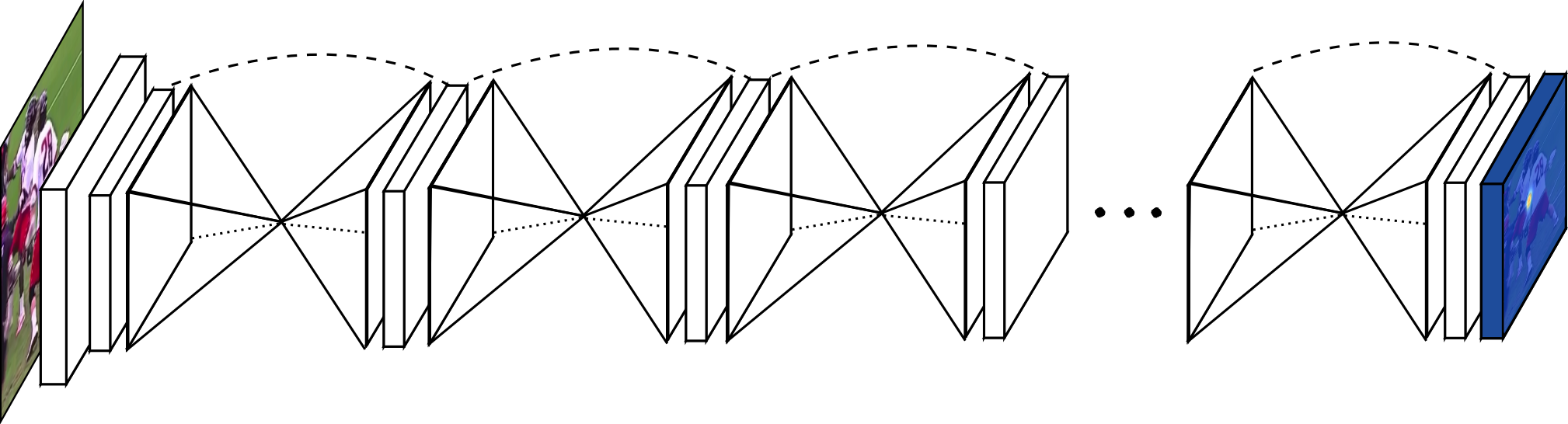}
\caption{Our network for pose estimation consists of multiple stacked
  hourglass modules which allow for repeated bottom-up, top-down
  inference.}
\label{fig:stacked-hg}
\end{figure}


\section{Introduction}

A key step toward understanding people in images and video is accurate
pose estimation. Given a single RGB image, we wish to determine the
precise pixel location of important keypoints of the body. Achieving
an understanding of a person's posture and limb articulation is
useful for higher level tasks like action recognition, and also
serves as a fundamental tool in fields such as human-computer
interaction and animation.

As a well established problem in vision, pose estimation has plagued
researchers with a variety of formidable challenges over the years. A
good pose estimation system must be robust to occlusion and severe
deformation, successful on rare and novel poses, and invariant to
changes in appearance due to factors like clothing and lighting. Early
work tackles such difficulties using robust image features and
sophisticated structured prediction \cite{sapp2013modec, felz08,
  pish13strong, bourdev2009poselets, johnson2011learning,
  ramanan2006learning, yang2013articulated, ferrari2008progressive,
  ladicky2013human}: the former is used to produce local
interpretations, whereas the latter is used to infer a globally
consistent pose.

This conventional pipeline, however, has been greatly reshaped by
convolutional neural networks (ConvNets) \cite{lecun1998gradient,
  krizhevsky2012imagenet, szegedy2015going, ioffe2015batch,
  he2015deep}, a main driver behind an explosive rise in performance
across many computer vision tasks. Recent pose estimation systems
\cite{tompson2014joint, tompson2015efficient, pish15deepcut,
  wei2016machines, carreira2015human, fan2015combining} have
universally adopted ConvNets as their main building block, largely
replacing hand-crafted features and graphical models; this strategy
has yielded drastic improvements on standard benchmarks
\cite{sapp2013modec, andriluka20142d, johnson2010clustered}.

We continue along this trajectory and introduce a novel ``stacked
hourglass'' network design for predicting human pose. The network
captures and consolidates information across all scales of the
image. We refer to the design as an hourglass based on our
visualization of the steps of pooling and subsequent upsampling used
to get the final output of the network. Like many convolutional
approaches that produce pixel-wise outputs, the hourglass network
pools down to a very low resolution, then upsamples and combines
features across multiple resolutions \cite{tompson2014joint,
  long2015fully}. On the other hand, the hourglass differs from prior
designs primarily in its more symmetric topology.

We expand on a single hourglass by consecutively placing multiple
hourglass modules together end-to-end. This allows for repeated
bottom-up, top-down inference across scales. In conjunction with the
use of intermediate supervision, repeated bidirectional inference is
critical to the network's final performance. The final network
architecture achieves a significant improvement on the
state-of-the-art for two standard pose estimation benchmarks (FLIC
\cite{sapp2013modec} and MPII Human Pose \cite{andriluka20142d}). On
MPII there is over a 2\% average accuracy improvement across all
joints, with as much as a 4-5\% improvement on more difficult joints
like the knees and ankles. \footnote{Code is available at
  \url{http://www-personal.umich.edu/~alnewell/pose}}


\section{Related Work}

With the introduction of ``DeepPose'' by Toshev et
al. \cite{toshev2014deeppose}, research on human pose estimation began
the shift from classic approaches~\cite{felz08, pish13strong,
  bourdev2009poselets, sapp2013modec, johnson2011learning,
  ramanan2006learning, yang2013articulated, ferrari2008progressive,
  ladicky2013human} to deep networks.  Toshev et al. use their network
to directly regress the x,y coordinates of joints. The work by Tompson
et al.~\cite{tompson2014joint} instead generates heatmaps by running
an image through multiple resolution banks in parallel to
simultaneously capture features at a variety of scales. Our network
design largely builds off of their work, exploring how to capture
information across scales and adapting their method for combining
features across different resolutions.

A critical feature of the method proposed by Tompson et
al. \cite{tompson2014joint} is the joint use of a ConvNet and a
graphical model. Their graphical model learns typical spatial
relationships between joints. Others have recently tackled this in
similar ways \cite{fan2015combining, chen2014articulated,
  pish15deepcut} with variations on how to approach unary score
generation and pairwise comparison of adjacent joints. Chen et
al. \cite{chen2014articulated} cluster detections into typical
orientations so that when their classifier makes predictions
additional information is available indicating the likely location of
a neighboring joint. We achieve superior performance without the use
of a graphical model or any explicit modeling of the human body.

\begin{figure}[t]
\centering
\includegraphics[width=\textwidth]{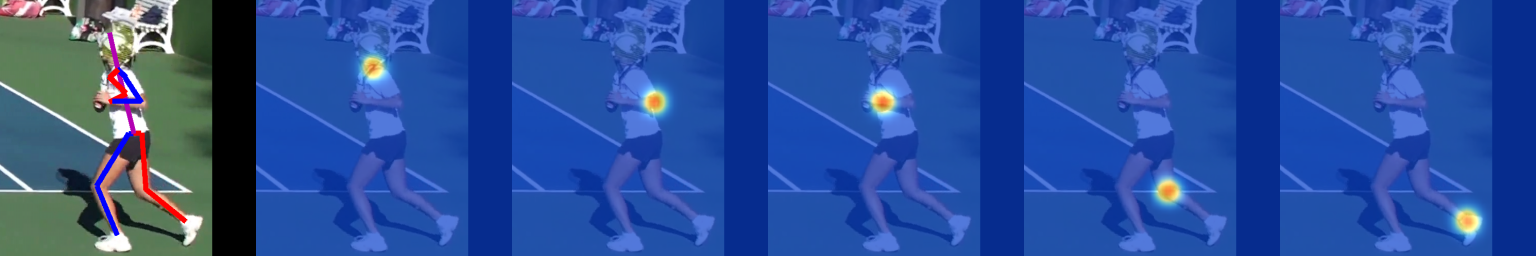}
\caption{Example output produced by our network. On the left we see
  the final pose estimate provided by the max activations across each
  heatmap. On the right we show sample heatmaps. (From left to right:
  neck, left elbow, left wrist, right knee, right ankle)}
\label{fig:exheatmaps}
\end{figure}

There are several examples of methods making successive predictions
for pose estimation. Carreira et al. \cite{carreira2015human} use what
they refer to as Iterative Error Feedback. A set of predictions is
included with the input, and each pass through the network further
refines these predictions. Their method requires multi-stage training
and the weights are shared across each iteration. Wei et
al. \cite{wei2016machines} build on the work of multi-stage pose
machines \cite{ramakrishna2014pose} but now with the use of ConvNets
for feature extraction. Given our use of intermediate supervision, our
work is similar in spirit to these methods, but our building block
(the hourglass module) is different. Hu \& Ramanan
\cite{hu2016bottomup} have an architecture more similar to ours that
can also be used for multiple stages of predictions, but their model
ties weights in the bottom-up and top-down portions of computation as
well as across iterations.

Tompson et al. build on their work in \cite{tompson2014joint} with a
cascade to refine predictions. This serves to increase efficency and
reduce memory usage of their method while improving localization
performance in the high precision range \cite{tompson2015efficient}.
One consideration is that for many failure cases a refinement of
position within a local window would not offer much improvement since
error cases often consist of either occluded or misattributed
limbs. For both situations, any further evaluation at a local scale
will not improve the prediction.
 
There are variations to the pose estimation problem which include the
use of additional features such as depth or motion cues.
\cite{jain2014modeep, shotton2013depth, pfister2015flowing} Also,
there is the more challenging task of simultaneous annotation of
multiple people \cite{chen2015parsing, pish15deepcut}. In addition,
there is work like that of Oliveira et al. \cite{oliveira2016deep}
that performs human part segmentation based on fully convolutional
networks~\cite{long2015fully}. Our work focuses solely on the task
of keypoint localization of a single person's pose from an RGB image.

Our hourglass module before stacking is closely connected to fully
convolutional networks \cite{long2015fully} and other designs that
process spatial information at multiple scales for dense
prediction~\cite{tompson2014joint, xie2015holistically,
  eigen2014depth, farabet2013learning, pinheiro2014recurrent,
  eigen2015predicting, mathieu2015deep, couprie2013indoor,
  bertasius2015deepedge, hariharan2015hypercolumns}. Xie et
al. \cite{xie2015holistically} give a summary of typical
architectures. Our hourglass module differs from these designs mainly
in its more symmetric distribution of capacity between bottom-up
processing (from high resolutions to low resolutions) and top-down
processing (from low resolutions to high resolutions). For example,
fully convolutional networks~\cite{long2015fully} and
holistically-nested architectures~\cite{xie2015holistically} are both
heavy in bottom-up processing but light in their top-down processing,
which consists only of a (weighted) merging of predictions across
multiple scales. Fully convolutional networks are also trained in
multiple stages.

The hourglass module before stacking is also related to conv-deconv
and encoder-decoder architectures~\cite{noh2015learning,
  zhao2015stacked, rematas2015deep, badrinarayanan2015segnet}.  Noh et
al. \cite{noh2015learning} use the conv-deconv architecture to do
semantic segmentation, Rematas et al. \cite{rematas2015deep} use it to
predict reflectance maps of objects. Zhao et
al. \cite{zhao2015stacked} develop a unified framework for supervised,
unsupervised and semi-supervised learning by adding a reconstruction
loss. Yang et al. \cite{yang2015nips} employ an encoder-decoder
architecture without skip connections for image generation. Rasmus et
al. \cite{rasmus2015semi} propose a denoising auto-encoder with
special, ``modulated'' skip connections for
unsupervised/semi-supervised feature learning.  The symmetric topology
of these networks is similar, but the nature of the operations is
quite different in that we do not use unpooling or deconv
layers. Instead, we rely on simple nearest neighbor upsampling and
skip connections for top-down processing. Another major difference of
our work is that we perform repeated bottom-up, top-down inference by
stacking multiple hourglasses.


\begin{figure}[t]
\centering
\includegraphics[width=0.75\textwidth]{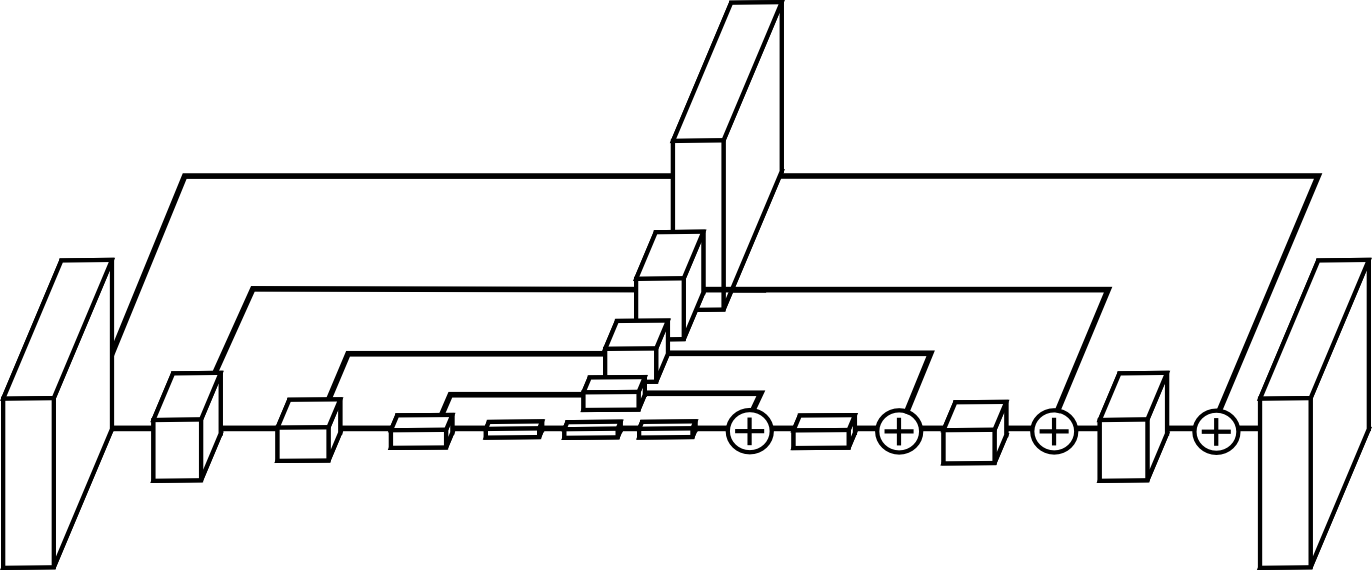}
\caption{An illustration of a single ``hourglass'' module. Each box in
  the figure corresponds to a residual module as seen in Figure
  \ref{fig:resid}. The number of features is consistent across the
  whole hourglass.}
\label{fig:single-hg}
\end{figure}

\section{Network Architecture}

\subsection{Hourglass Design}

The design of the hourglass is motivated by the need to capture
information at every scale. While local evidence is essential for
identifying features like faces and hands, a final pose estimate
requires a coherent understanding of the full body. The person's
orientation, the arrangement of their limbs, and the relationships of
adjacent joints are among the many cues that are best recognized at
different scales in the image. The hourglass is a simple, minimal
design that has the capacity to capture all of these features and
bring them together to output pixel-wise predictions.

The network must have some mechanism to effectively process and
consolidate features across scales. Some approaches tackle this with
the use of separate pipelines that process the image independently at
multiple resolutions and combine features later on in the network
\cite{tompson2014joint, wei2016machines}. Instead, we choose to use a
single pipeline with skip layers to preserve spatial information at
each resolution. The network reaches its lowest resolution at 4x4
pixels allowing smaller spatial filters to be applied that compare
features across the entire space of the image.

The hourglass is set up as follows: Convolutional and max pooling
layers are used to process features down to a very low resolution. At
each max pooling step, the network branches off and applies more
convolutions at the original pre-pooled resolution. After reaching the
lowest resolution, the network begins the top-down sequence of
upsampling and combination of features across scales. To bring
together information across two adjacent resolutions, we follow the
process described by Tompson et al. \cite{tompson2014joint} and do
nearest neighbor upsampling of the lower resolution followed by an
elementwise addition of the two sets of features. The topology of the
hourglass is symmetric, so for every layer present on the way down
there is a corresponding layer going up.

After reaching the output resolution of the network, two consecutive
rounds of 1x1 convolutions are applied to produce the final network
predictions. The output of the network is a set of heatmaps where for
a given heatmap the network predicts the probability of a joint's
presence at each and every pixel. The full module (excluding the final
1x1 layers) is illustrated in Figure \ref{fig:single-hg}.

\subsection{Layer Implementation}

While maintaining the overall hourglass shape, there is still some
flexibility in the specific implementation of layers. Different
choices can have a moderate impact on the final performance and
training of the network. We explore several options for layer design
in our network. Recent work has shown the value of reduction steps
with 1x1 convolutions, as well as the benefits of using consecutive
smaller filters to capture a larger spatial context.\cite{he2015deep,
  szegedy2015going} For example, one can replace a 5x5 filter with two
separate 3x3 filters. We tested our overall network design, swapping
in different layer modules based off of these insights. We experienced
an increase in network performance after switching from standard
convolutional layers with large filters and no reduction steps to
newer methods like the residual learning modules presented by He et
al. \cite{he2015deep} and ``Inception''-based designs
\cite{szegedy2015going}. After the initial performance improvement
with these types of designs, various additional explorations and
modifications to the layers did little to further boost performance or
training time.

\begin{figure}[t]
\centering
\includegraphics[width=.4\textwidth]{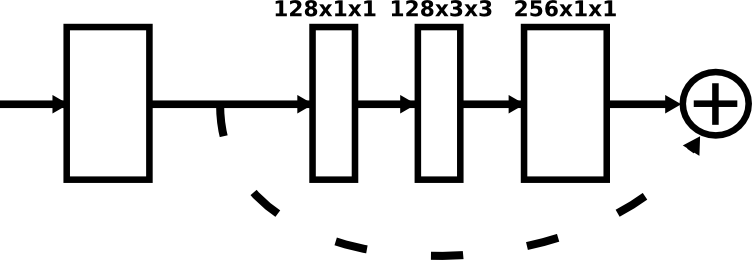}
\hspace{.1in}
\includegraphics[width=.55\textwidth]{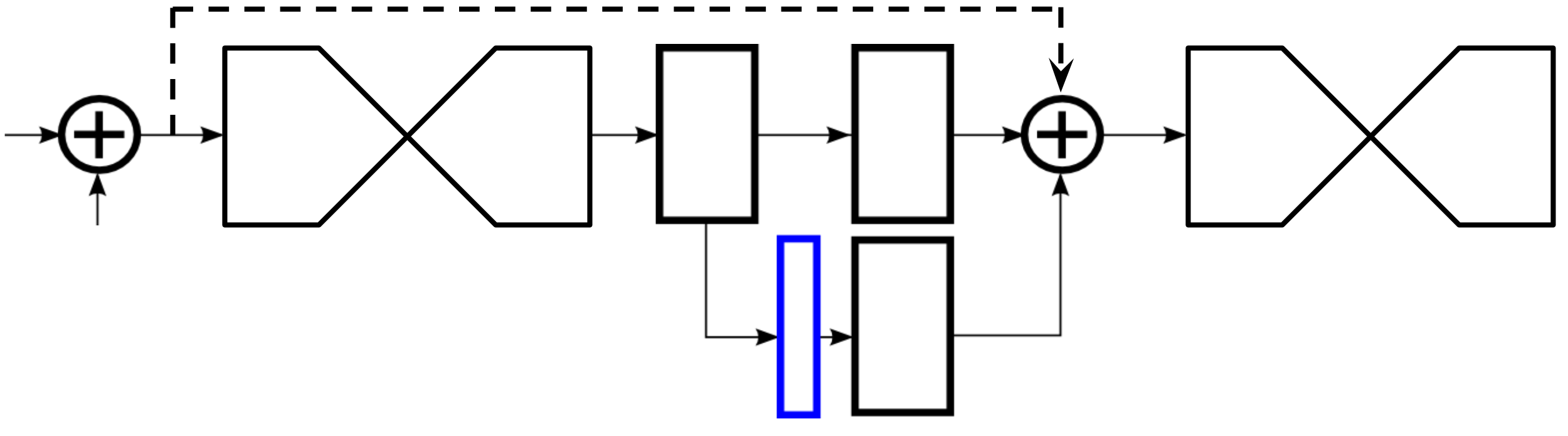}
\caption{\textbf{Left:} Residual Module \cite{he2015deep} that we use
  throughout our network. \textbf{Right:} Illustration of the
  intermediate supervision process. The network splits and produces a set of
  heatmaps (outlined in blue) where a loss can be applied. A 1x1
  convolution remaps the heatmaps to match the number of channels of
  the intermediate features. These are added together along with the
  features from the preceding hourglass.}
\label{fig:resid}
\end{figure}

Our final design makes extensive use of residual modules. Filters
greater than 3x3 are never used, and the bottlenecking restricts the
total number of parameters at each layer curtailing total memory
usage. The module used in our network is shown in Figure
\ref{fig:resid}. To put this into the context of the full network
design, each box in Figure \ref{fig:single-hg} represents a single
residual module.

Operating at the full input resolution of 256x256 requires a
significant amount of GPU memory, so the highest resolution of the
hourglass (and thus the final output resolution) is 64x64. This does
not affect the network's ability to produce precise joint
predictions. The full network starts with a 7x7 convolutional layer
with stride 2, followed by a residual module and a round of max
pooling to bring the resolution down from 256 to 64. Two subsequent
residual modules precede the hourglass shown in Figure
\ref{fig:single-hg}. Across the entire hourglass all residual modules
output 256 features.

\subsection{Stacked Hourglass with Intermediate Supervision}

We take our network architecture further by stacking multiple
hourglasses end-to-end, feeding the output of one as input into the
next. This provides the network with a mechanism for repeated
bottom-up, top-down inference allowing for reevaluation of initial
estimates and features across the whole image. The key to this
approach is the prediction of intermediate heatmaps upon which we can
apply a loss. Predictions are generated after passing through each
hourglass where the network has had an opportunity to process features
at both local and global contexts. Subsequent hourglass modules allow
these high level features to be processed again to further evaluate
and reassess higher order spatial relationships. This is similar to
other pose estimations methods that have demonstrated strong
performance with multiple iterative stages and intermediate
supervision \cite{carreira2015human,wei2016machines,pfister2015flowing}.

Consider the limits of applying intermediate supervision with only the
use of a single hourglass module. What would be an appropriate place
in the pipeline to generate an initial set of predictions? Most higher
order features are present only at lower resolutions except at the
very end when upsampling occurs. If supervision is provided after the
network does upsampling then there is no way for these features to be
reevaluated relative to each other in a larger global context. If we
want the network to best refine predictions, these predictions cannot
be exclusively evaluated at a local scale. The relationship to other
joint predictions as well as the general context and understanding of
the full image is crucial. Applying supervision earlier in
the pipeline before pooling is a possibility, but at this point the
features at a given pixel are the result of processing a relatively
local receptive field and are thus ignorant of critical global cues.

Repeated bottom-up, top-down inference with stacked hourglasses
alleviates these concerns. Local and global cues are integrated within
each hourglass module, and asking the network to produce early
predictions requires it to have a high-level understanding of the
image while only partway through the full network. Subsequent stages
of bottom-up, top-down processing allow for a deeper reconsideration
of these features.

This approach for going back and forth between scales is particularly
important because preserving the spatial location of features is
essential to do the final localization step. The precise position of a
joint is an indispensable cue for other decisions being made by the
network. With a structured problem like pose estimation, the output is
an interplay of many different features that should come together to
form a coherent understanding of the scene. Contradicting evidence and
anatomic impossiblity are big giveaways that somewhere along the line
a mistake was made, and by going back and forth the network can
maintain precise local information while considering and then
reconsidering the overall coherence of the features.

We reintegrate intermediate predictions back into the feature space by
mapping them to a larger number of channels with an additional 1x1
convolution. These are added back to the intermediate features from
the hourglass along with the features output from the previous
hourglass stage (visualized in Figure \ref{fig:resid}). The resulting
output serves directly as the input for the following hourglass module
which generates another set of predictions. In the final network
design, eight hourglasses are used. It is important to note that
weights are not shared across hourglass modules, and a loss is applied
to the predictions of all hourglasses using the same ground truth. The
details for the loss and ground truth are described below.

\subsection{Training Details}

We evaluate our network on two benchmark datasets, FLIC
\cite{sapp2013modec} and MPII Human Pose \cite{andriluka20142d}. FLIC
is composed of 5003 images (3987 training, 1016 testing) taken from
films. The images are annotated on the upper body with most
figures facing the camera straight on. MPII Human Pose consists of
around 25k images with annotations for multiple people providing 40k
annotated samples (28k training, 11k testing). The test annotations
are not provided so in all of our experiments we train on a subset of
training images while evaluating on a heldout validation set of around
3000 samples. MPII consists of images taken from a wide range of human
activities with a challenging array of widely articulated full-body
poses.

\begin{figure}[t]
\centering
\includegraphics[width=\textwidth]{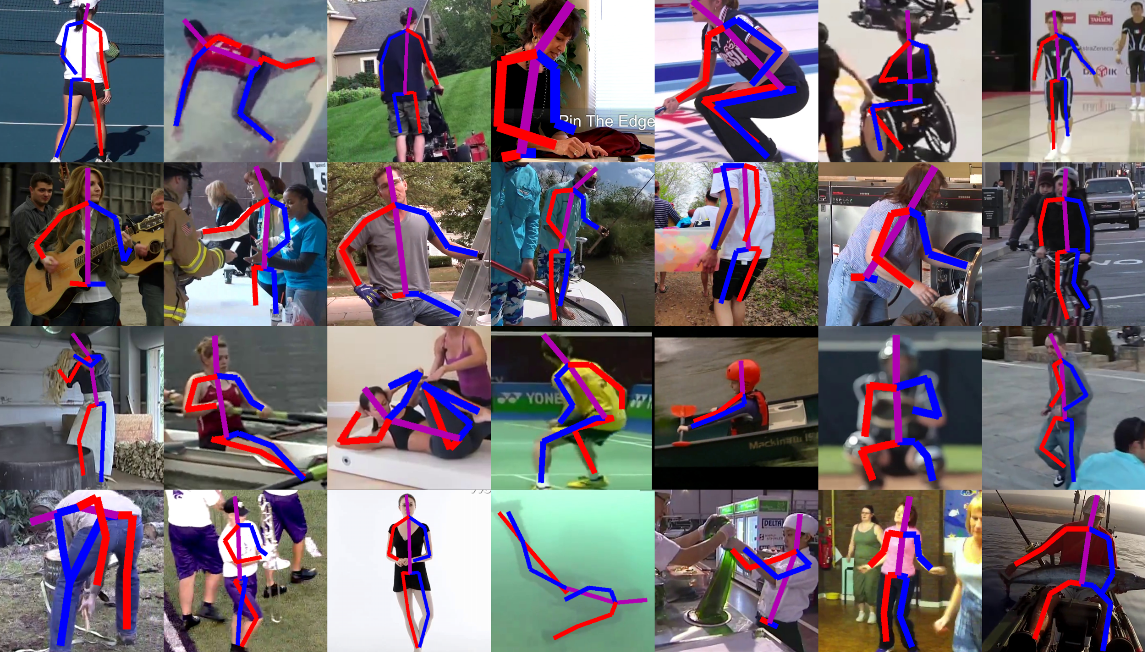}
\caption{Example output on MPII's test set.}
\label{fig:pos-examples}
\end{figure}

There are often multiple people visible in a given input image, but
without a graphical model or other postprocessing step the image must
convey all necessary information for the network to determine which
person deserves the annotation. We deal with this by training the
network to exclusively annotate the person in the direct center. This
is done in FLIC by centering along the x-axis according to the
torsobox annotation - no vertical adjustment or scale normalization is
done. For MPII, it is standard to utilize the scale and center
annotations provided with all images. For each sample, these values
are used to crop the image around the target person. All input images
are then resized to 256x256 pixels. We do data augmentation that
includes rotation (+/- 30 degrees), and scaling (.75-1.25). We avoid
translation augmentation of the image since location of the target
person is the critical cue determining who should be annotated by the
network.

The network is trained using Torch7 \cite{torch7} and for optimization
we use rmsprop \cite{rmsprop} with a learning rate of 2.5e-4. Training
takes about 3 days on a 12 GB NVIDIA TitanX GPU. We drop the learning
rate once by a factor of 5 after validation accuracy plateaus. Batch
normalization \cite{ioffe2015batch} is also used to improve
training. A single forward pass of the network takes 75 ms. For
generating final test predictions we run both the original input
and a flipped version of the image through the network and average the
heatmaps together (accounting for a 1\% average improvement on
validation). The final prediction of the network is the max
activating location of the heatmap for a given joint.

The same technique as Tompson et al. \cite{tompson2014joint} is used
for supervision. A Mean-Squared Error (MSE) loss is applied comparing
the predicted heatmap to a ground-truth heatmap consisting of a 2D
gaussian (with standard deviation of 1 px) centered on the joint
location. To improve performance at high precision thresholds the
prediction is offset by a quarter of a pixel in the direction of its
next highest neighbor before transforming back to the original
coordinate space of the image. In MPII Human Pose, some joints do not
have a corresponding ground truth annotation. In these cases the joint
is either truncated or severely occluded, so for supervision a ground
truth heatmap of all zeros is provided.


\section{Results}

\begin{figure}[t]
\begin{floatrow}
\ffigbox{
  \includegraphics[trim={.4in .1in 0 0},width=.5\textwidth,height=1.2in,clip]{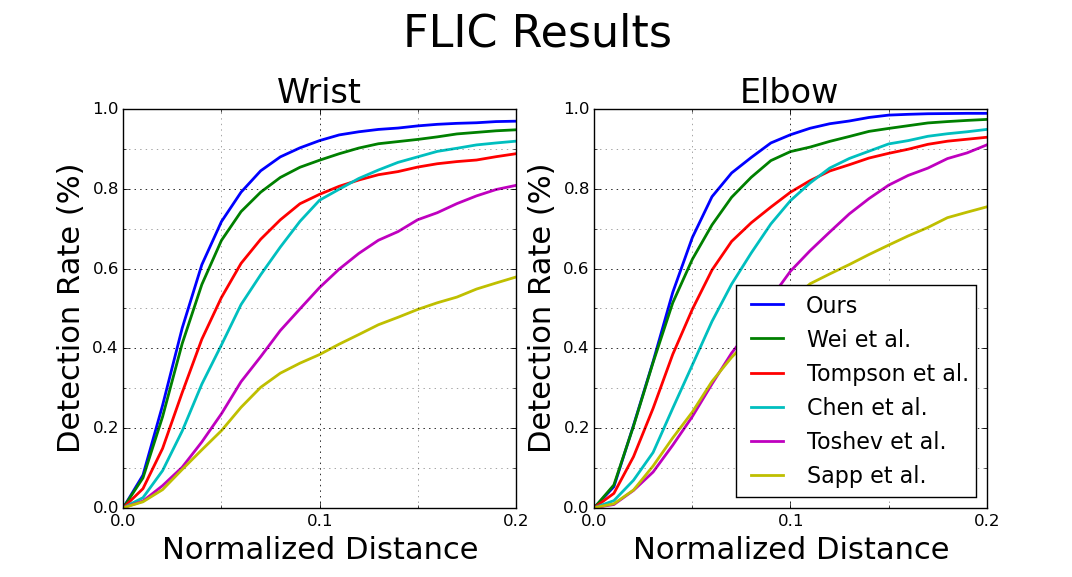}
}{\caption{PCK comparison on FLIC}
  \label{fig:flic-pck}}
\captabbox{
  \begin{tabular}{l||*{7}{c}|r} \hline
    & Elbow & Wrist \\ \hline
    Sapp et al. \cite{sapp2013modec} & 76.5  & 59.1  \\
    Toshev et al. \cite{toshev2014deeppose} & 92.3  & 82.0  \\
    Tompson et al. \cite{tompson2015efficient} & 93.1  & 89.0 \\
    Chen et al. \cite{chen2014articulated} & 95.3  & 92.4 \\
    Wei et al. \cite {wei2016machines} & 97.6 & 95.0 \\ \hline
    Our model & \textbf{99.0}  & \textbf{97.0}  \\ \hline
  \end{tabular}
}{\caption{FLIC results (PCK@0.2)}
  \label{table:flic-pck2}}
\end{floatrow}
\captabfig{
\includegraphics[trim={0 .2in 0 -.3in},width=\textwidth,clip]{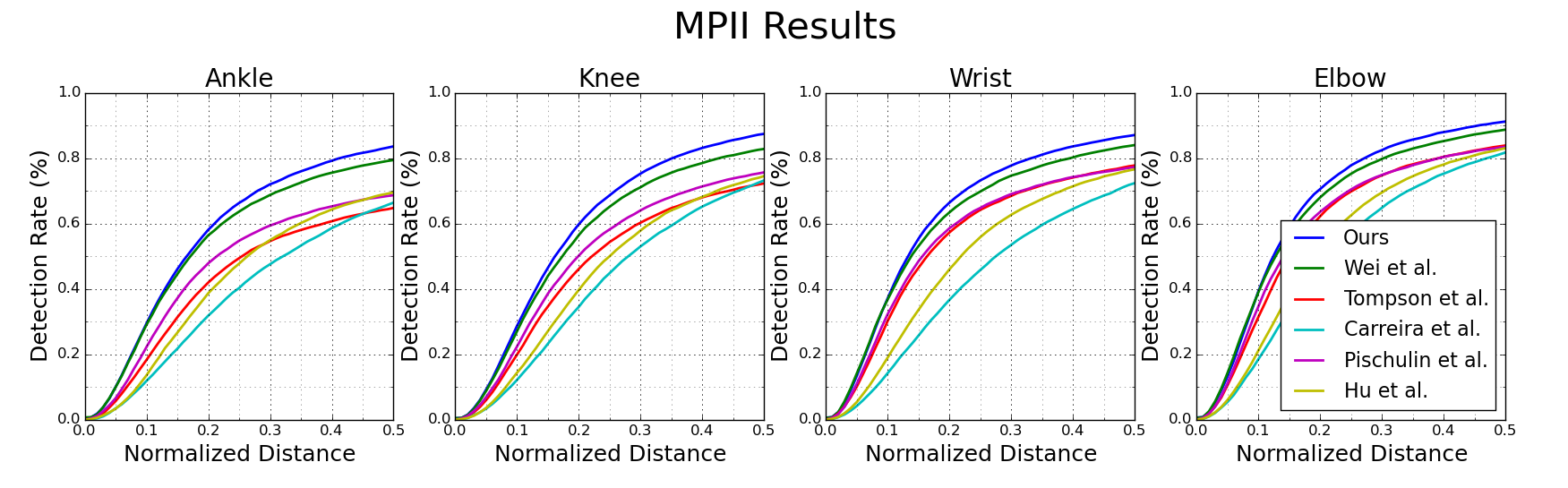}
}{\caption{PCKh comparison on MPII}
\label{fig:mpii-pckh}}
\captabbox{
\begin{tabular}{l||*{7}{c}|r}
  \hline
  & Head & Shoulder & Elbow & Wrist & Hip & Knee  & Ankle & Total \\ \hline
  Tompson et al. \cite{tompson2015efficient}, CVPR'15&96.1 &91.9 &83.9 &77.8 &80.9 &72.3 &64.8 &82.0\\
  Carreira et al. \cite{carreira2015human}, CVPR'16 &95.7 &91.7 &81.7 &72.4 &82.8 &73.2 &66.4 &81.3\\
  Pishchulin et al. \cite{pish15deepcut}, CVPR'16 &94.1 &90.2 &83.4 &77.3 &82.6 &75.7 & 68.6 & 82.4 \\
  Hu et al. \cite{hu2016bottomup}, CVPR'16 &95.0 &91.6 &83.0 &76.6 &81.9 &74.5 &69.5 &82.4 \\
  Wei et al. \cite{wei2016machines}, CVPR'16 &97.8 &95.0 &88.7 &84.0 &88.4 &82.8 &79.4 &88.5 \\ \hline
  Our model& \textbf{98.2} &\textbf{96.3} &\textbf{91.2} &\textbf{87.1}
           & \textbf{90.1} &\textbf{87.4} & \textbf{83.6} & \textbf{90.9} \\ \hline 
\end{tabular}
}{\caption{Results on MPII Human Pose (PCKh@0.5)}
\label{table:mpii-pckh5}}
\end{figure}

\subsection{Evaluation}

Evaluation is done using the standard Percentage of Correct Keypoints
(PCK) metric which reports the percentage of detections that fall
within a normalized distance of the ground truth. For FLIC, distance
is normalized by torso size, and for MPII, by a fraction of the head
size (referred to as PCKh).

\textbf{FLIC:} Results can be seen in Figure \ref{fig:flic-pck} and
Table \ref{table:flic-pck2}.  Our results on FLIC are very competitive
reaching 99\% PCK@0.2 accuracy on the elbow, and 97\% on the wrist. It
is important to note that these results are observer-centric, which is
consistent with how others have evaluated their output on
FLIC.

\textbf{MPII:} We achieve state-of-the-art results across all joints
on the MPII Human Pose dataset. All numbers can be seen in Table
\ref{table:mpii-pckh5} along with PCK curves in Figure
\ref{fig:mpii-pckh}. On difficult joints like the wrist, elbows,
knees, and ankles we improve upon the most recent state-of-the-art
results by an average of 3.5\% (PCKh@0.5) with an average error rate
of 12.8\% down from 16.3\%. The final elbow accuracy is 91.2\% and
wrist accuracy is 87.1\%. Example predictions made by the network on
MPII can be seen in Figure \ref{fig:pos-examples}.

\subsection{Ablation Experiments}

\begin{figure}[t]
\centering
\includegraphics[width=.55\textwidth]{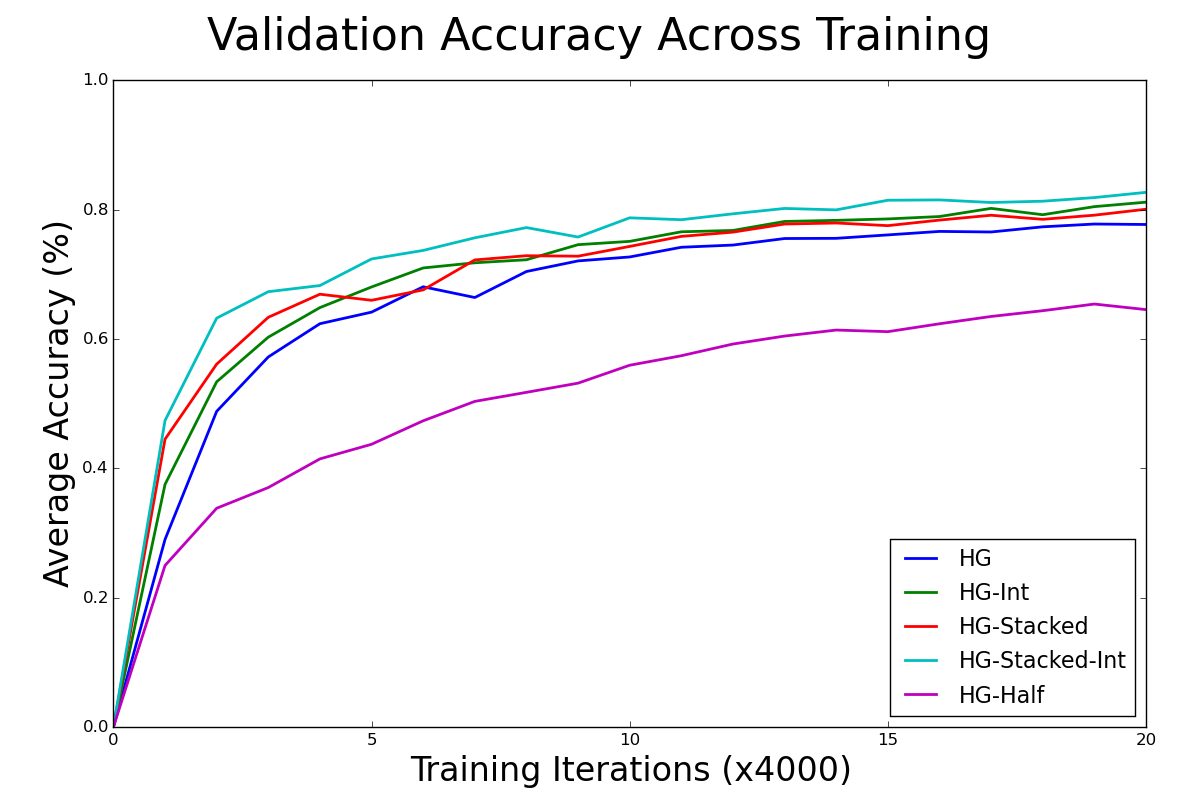}
\includegraphics[width=.4\textwidth]{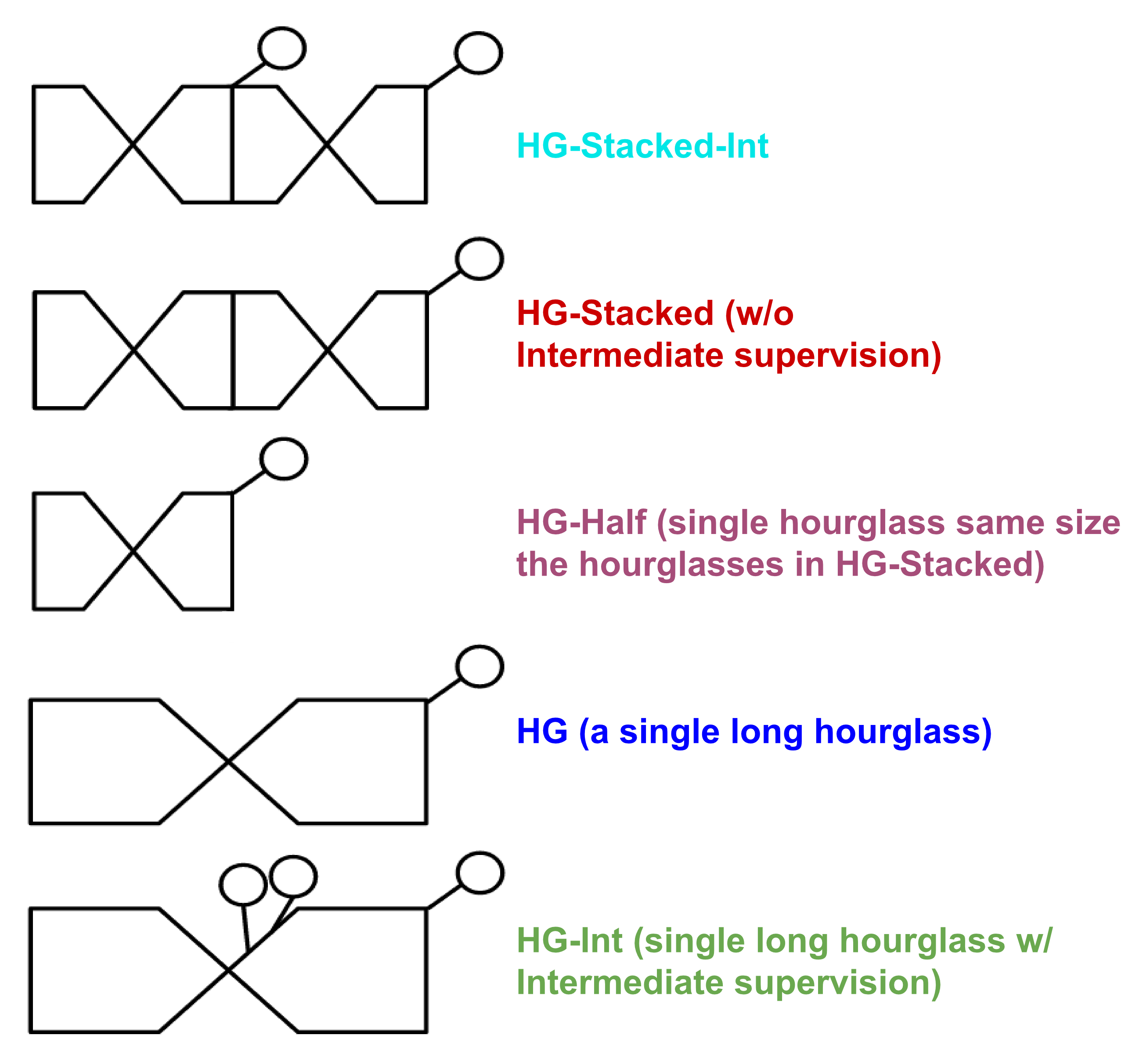}
\caption{ Comparison of validation accuracy as training
  progresses. The accuracy is averaged across the wrists, elbows,
  knees, and ankles. The different network designs are illustrated on
  the right, the circle is used to indicate where a loss is applied}
\label{fig:ablation}
\end{figure}

We explore two main design choices in this work: the effect of
stacking hourglass modules together, and the impact of intermediate
supervision. These are not mutually independent as we are limited in
how we can apply intermediate supervision depending on the overall
architectural design. Applied separately, each has a positive impact
on performance, and together we see a further improvements to training
speed and in the end, final pose estimation performance. We look at
the rate of training of a few different network designs. The results
of which can be seen in Figure \ref{fig:ablation} which shows average
accuracy on the validation set as training progresses. The accuracy
metric considers all joints excluding those associated with the head
and torso to allow for easier differentiation across experiments.

\begin{figure}[t]
\centering
\includegraphics[width=.45\textwidth]{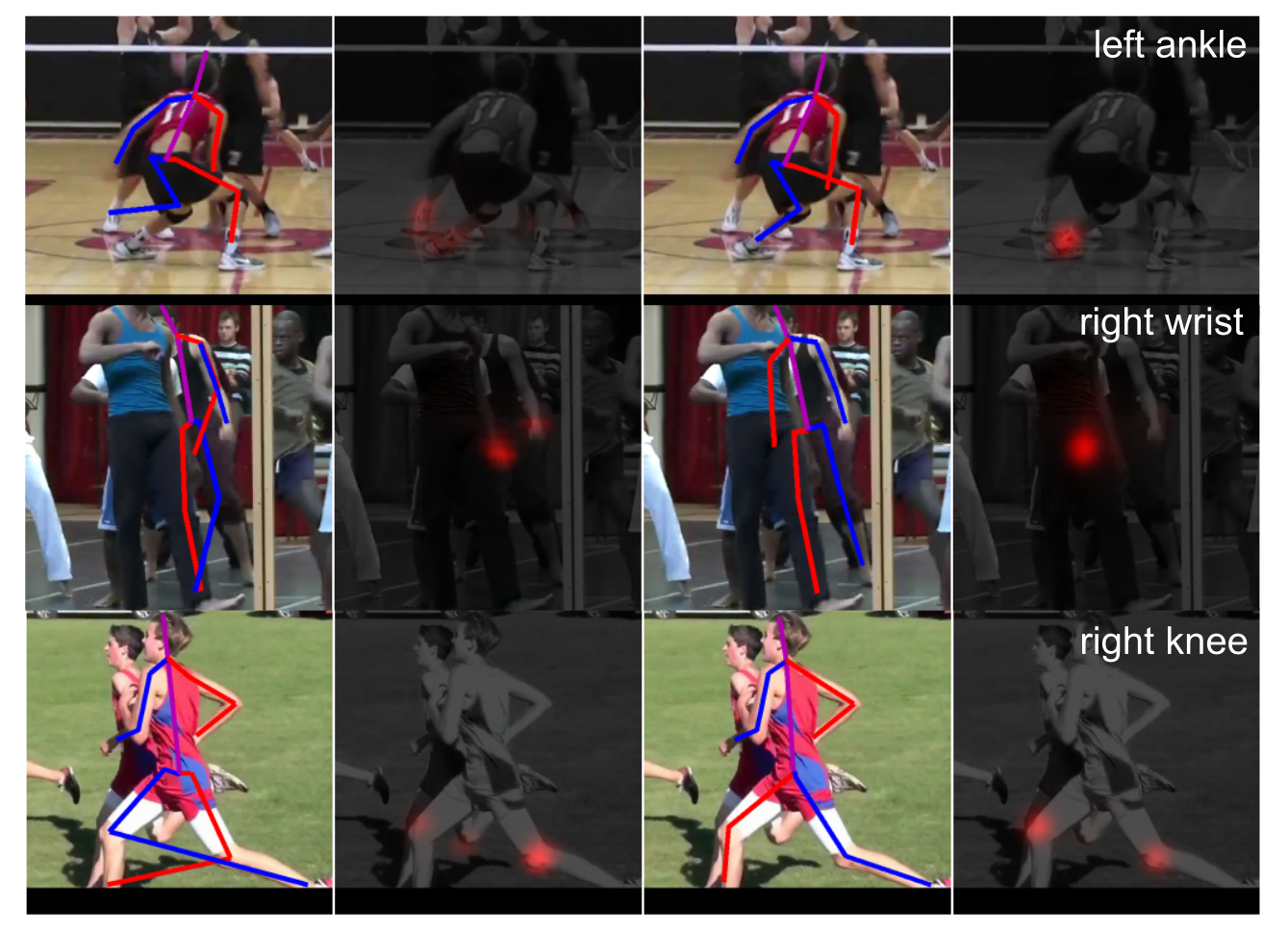}
\includegraphics[width=.53\textwidth]{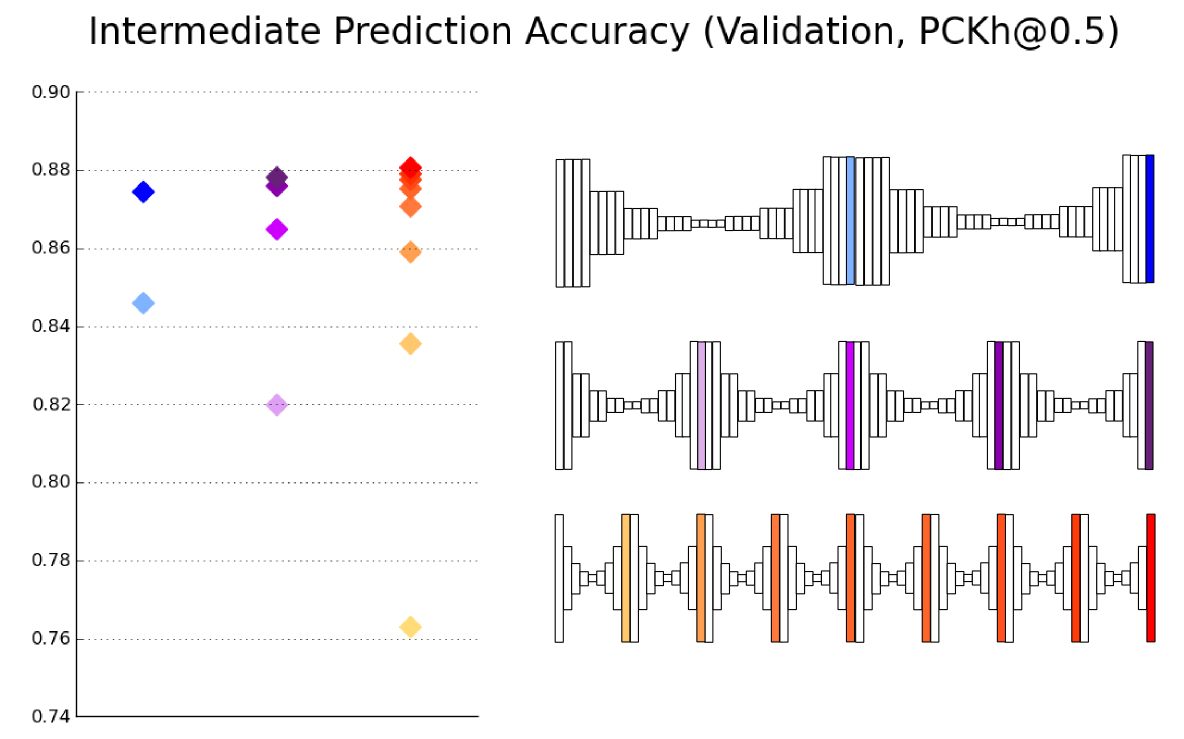}
\caption{\textbf{Left:} Example validation images illustrating the
  change in predictions from an intermediate stage (second hourglass)
  (left) to final predictions (eighth hourglass)
  (right). \textbf{Right:} Validation accuracy at intermediate stages
  of the network compared across different stacking arrangements.}
\label{fig:inter}
\end{figure}

First, to explore the effect of the stacked hourglass design we must
demonstrate that the change in performance is a function of the
architecture shape and not attributed to an increase in capacity with
a larger, deeper network. To make this comparison, we work from a
baseline network consisting of eight hourglass modules stacked
together. Each hourglass has a single residual module at each
resolution as in Figure \ref{fig:single-hg}. We can shuffle these
layers around for various network arrangements. A decrease in the
number of hourglasses would result in an increase in the capacity of
each hourglass. For example, a corresponding network could stack four
hourglasses and have two consecutive residual modules at each
resolution (or two hourglasses and four residual modules). This is
illustrated in Figure \ref{fig:inter}. All networks share the same
number of parameters and layers, though a slight difference is
introduced when more intermediate supervision is applied.

To see the effect of these choices we first compare a two-stacked
network with four residual modules at each stage in the hourglass, and
a single hourglass but with eight residual modules instead. In Figure
\ref{fig:ablation} these are referred to as HG-Stacked and HG
respectively. A modest improvement in training can be seen when using
the stacked design despite having approximately the same number of
layers and parameters. Next, we consider the impact of intermediate
supervision. For the two-stack network we follow the procedure
described in the paper to apply supervision. Applying this same idea
with a single hourglass is nontrivial since higher order global
features are present only at lower resolutions, and the features
across scales are not combined until late in the pipeline. We explore
applying supervision at various points in the network, for example
either before or after pooling and at various resolutions. The best
performing method is shown as HG-Int in Figure \ref{fig:ablation} with
intermediate supervision applied after upsampling at the next two
highest resolutions before the final output resolution. This
supervision does offer an improvement to performance, but not enough
to surpass the improvement when stacking is included (HG-Stacked-Int).

In Figure \ref{fig:inter} we compare the validation accuracy of 2-,
4-, and 8-stack models that share approximately the same number of
parameters, and include the accuracy of their intermediate
predictions. There is a modest improvement in final performance for
each successive increase in stacking from 87.4\% to 87.8\% to
88.1\%. The effect is more notable at intermediate stages. For
example, halfway through each network the corresponding accuracies of
the intermediate predictions are: 84.6\%, 86.5\%, and 87.1\%. Note
that the accuracy halfway through the 8-stack network is just short of
the final accuracy of the 2-stack network.

It is interesting to observe the mistakes made early and corrected
later on by the network. A few examples are visualized in Figure
\ref{fig:inter}. Common mistakes show up like a mix up of other
people's joints, or misattribution of left and right. For the running
figure, it is apparent from the final heatmap that the decision
between left and right is still a bit ambiguous for the network. Given
the appearance of the image, the confusion is justified. One case
worth noting is the middle example where the network initially
activates on the visible wrists in the image. Upon further processing
the heatmap does not activate at all on the original locations, instead
choosing a reasonable position for the occluded wrist.


\section{Further Analysis}

\subsection{Multiple People}

\begin{figure}[t]
  \centering {\includegraphics[width=\textwidth]{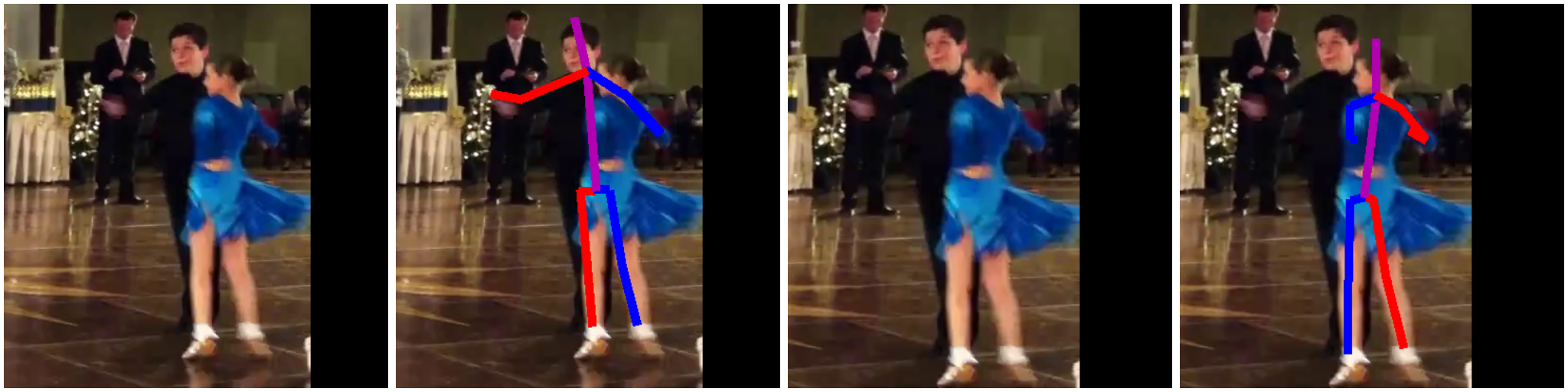}}
  \caption{The difference made by a slight translation and change of
    scale of the input image. The network determines who to generate
    an annotation for based on the central figure. The scaling and
    shift right of the input image is enough for the network to
    switch its predictions.}
  \label{fig:multi}
\end{figure}

The issue of coherence becomes especially important when there are
multiple people in an image. The network has to decide who to
annotate, but there are limited options for communicating who exactly
deserves the annotation. For the purposes of this work, the only
signal provided is the centering and scaling of the target person
trusting that the input will be clear enough to parse. Unfortunately,
this occasionally leads to ambiguous situations when people are very
close together or even overlapping as seen in Figure
\ref{fig:multi}. Since we are training a system to generate pose
predictions for a single person, the ideal output in an ambiguous
situation would demonstrate a commitment to the joints of just one
figure. Even if the predictions are lower quality, this would show a
deeper understanding of the task at hand. Estimating a location for
the wrist with a disregard for whom the wrist may belong is not
desired behavior from a pose estimation system.

The results in Figure \ref{fig:multi} are from an MPII test image. The
network must produce predictions for both the boy and girl, and to do
so, their respective center and scale annotations are provided. Using
those values to crop input images for the network result in the first
and third images of the figure. The center annotations for the two
dancers are off by just 26 pixels in a 720x1280 image. Qualitatively,
the most perceptible difference between the two input images is the
change in scale. This difference is sufficient for the network to
change its estimate entirely and predict the annotations for the
correct figure.

A more comprehensive management of annotations for multiple people is
out of the scope of this work. Many of the system's failure cases are
a result of confusing the joints of multiple people, but it is
promising that in many examples with severe overlap of figures the
network will appropriately pick out a single figure to annotate. 

\subsection{Occlusion}

Occlusion performance can be difficult to assess as it often falls
into two distinct categories. The first consists of cases where a
joint is not visible but its position is apparent given the context of
the image. MPII generally provides ground truth locations for these
joints, and an additional annotation indicates their lack of
visibility. The second situation, on the other hand, occurs when there
is absolutely no information about where a particular joint might
be. For example, images where only the upper half of the person's body
is visible. In MPII these joints will not have a ground truth
annotation associated with them.

\begin{figure}[t]
\centering
\includegraphics[width=\textwidth]{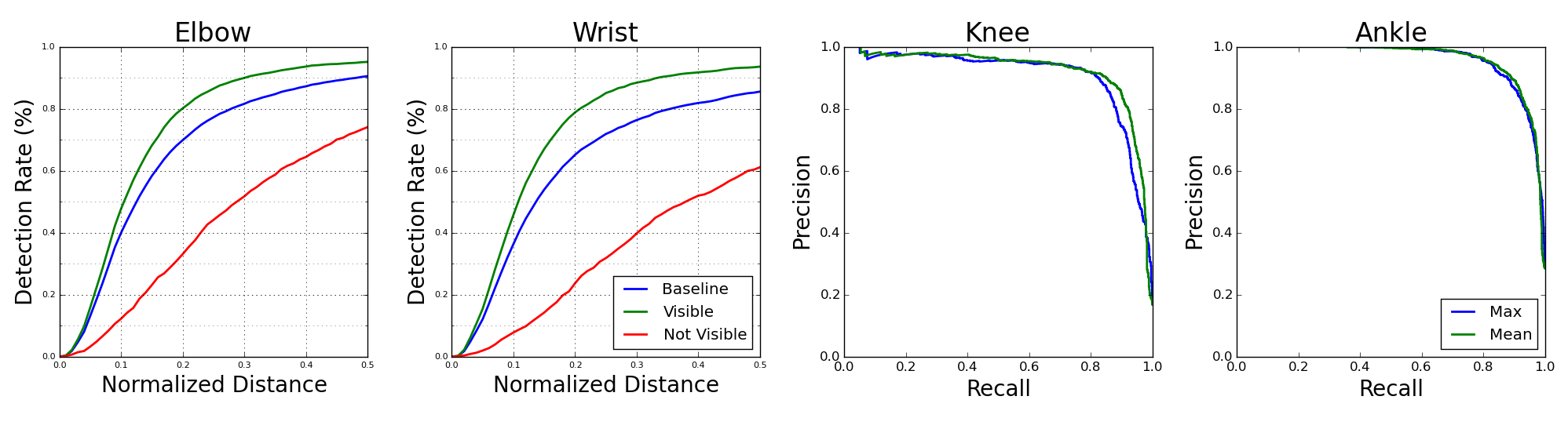}
\caption{\textbf{Left:} PCKh curves on validation comparing
  performance when exclusively considering joints that are visible (or
  not). \textbf{Right:} Precision recall curves showing the accuracy
  of predicting whether an annotation is present for a joint when
  thresholding on either the mean or max activation of a heatmap.}
\label{fig:occlusion}
\end{figure}

Our system makes no use of the additional visibility annotations, but
we can still take a look at the impact of visibility on
performance. About 75\% of the elbows and wrists with annotations are
labeled visible in our held-out validation set. In Figure
\ref{fig:occlusion}, we compare performance averaged across the whole
validation set with performance on the three-quarters of joints that
are visible and performance on the remaining quarter that are
not. While only considering visible joints, wrist accuracy goes up to
93.6\% from 85.5\% (validation performance is slightly worse than test
set performance of 87.1\%). On the other hand, performance on
exclusively occluded joints is 61.1\%. For the elbow, accuracy goes
from a baseline of 90.5\% to 95.1\% for visible joints and down to
74.0\% for occluded joints. Occlusion is clearly a significant
challenge, but the network still makes strong estimates in most
cases. In many examples, the network prediction and ground-truth
annotation may not agree while both residing in valid locations, and
the ambiguity of the image means there is no way to determine which
one is truly correct.

We also consider the more extreme case where a joint may be severely
occluded or truncated and therefore have no annotation at all. The PCK
metric used when evaluating pose estimation systems does not reflect
how well these situations are recognized by the network. If there is
no ground truth annotation provided for a joint it is impossible to
assess the quality of the prediction made by the system, so it is not
counted towards the final reported PCK value. Because of this, there
is no harm in generating predictions for all joints even though the
predictions for completely occluded or truncated joints will make no
sense. For use in a real system, a degree of metaknowledge is
essential, and the understanding that no good prediction can be made
on a particular joint is very important. We observe that our network
gives consistent and accurate predictions of whether or not a ground
truth annotation is available for a joint.

We consider the ankle and knee for this analysis since these are
occluded most often. Lower limbs are frequently cropped from images,
and if we were to always visualize all joint predictions of our
network, example pose figures would look unacceptable given the
nonsensical lower body predictions made in these situations. For a
simple way to filter out these cases we examine how well one can
determine the presence of an annotation for a joint given the
corresponding heatmap activation. We consider thresholding on either
the maximum value of the heatmap or its mean. The corresponding
precision-recall curves can be seen in Figure \ref{fig:occlusion}. We
find that based solely off of the mean activation of a heatmap it is
possible to correctly assess the presence of an annotation for the
knee with an AUC of 92.1\% and an annotation for the ankle with
an AUC of 96.0\%. This was done on a validation set of 2958
samples of which 16.1\% of possible knees and 28.4\% of possible
ankles do not have a ground truth annotation. This is a promising
result demonstrating that the heatmap serves as a useful signal
indicating cases of truncation and severe occlusion in images.


\section{Conclusion}

We demonstrate the effectiveness of a stacked hourglass network for
producing human pose estimates. The network handles a diverse and
challenging set of poses with a simple mechanism for reevaluation and
assessment of initial predictions. Intermediate supervision is
critical for training the network, working best in the context of
stacked hourglass modules. There still exist difficult cases not
handled perfectly by the network, but overall our system shows robust
performance to a variety of challenges including heavy occlusion and
multiple people in close proximity.

\bibliographystyle{splncs}
\bibliography{egbib}
\end{document}